\documentclass{article}

     \PassOptionsToPackage{numbers, compress}{natbib}

\usepackage[final]{neurips_2024}

\usepackage[utf8]{inputenc} 
\usepackage[T1]{fontenc}    
\usepackage{hyperref}       
\usepackage{url}            
\usepackage{booktabs}       
\usepackage{amsfonts}       
\usepackage{nicefrac}       
\usepackage{microtype}      
\usepackage{xcolor}         
\usepackage[utf8]{inputenc} 
\usepackage[T1]{fontenc}    
\usepackage{xcolor}

\usepackage{algorithm}
\usepackage{amsfonts}       
\usepackage{amsmath}
\usepackage{amssymb}
\usepackage{amsthm}
\usepackage{array}
\usepackage{bm}
\usepackage{booktabs}       
\usepackage{caption}
\usepackage{colortbl}
\usepackage{empheq}
\usepackage{enumitem}
\usepackage{etoolbox}
\usepackage{float}
\usepackage{makecell}
\usepackage{mathrsfs}
\usepackage{mdframed}
\usepackage{microtype}      
\usepackage{multirow}
\usepackage{multicol}
\usepackage{nccmath}
\usepackage{nicefrac}       
\usepackage[noend]{algorithmic}
\usepackage{pifont}
\usepackage{ragged2e}
\usepackage{rotating}
\usepackage{subcaption}
\usepackage{tabularx}
\usepackage{tikz}
\usetikzlibrary{tikzmark}
\usepackage{times}
\usepackage{url}            
\usepackage{wrapfig}
\usepackage{color, colortbl}
\usepackage{xspace}
\usepackage{thm-restate}
\usepackage{xfrac}

\newcounter{rowcntr}[table]
\renewcommand{\therowcntr}{\arabic{chapter}.\the\numexpr\arabic{table}+1.\arabic{rowcntr}}

\AtBeginEnvironment{tabular}{\setcounter{rowcntr}{0}}
\newcolumntype{H}{>{\setbox0=\hbox\bgroup}c<{\egroup}@{}}

\newcommand{\PreserveBackslash}[1]{\let\temp=\\#1\let\\=\temp}
\newcolumntype{C}[1]{>{\centering\arraybackslash}m{#1}}
\newcolumntype{R}[1]{>{\raggedleft\arraybackslash}m{#1}}
\newcolumntype{L}[1]{>{\raggedright\arraybackslash}m{#1}}

\usepackage[capitalize]{cleveref}
\usepackage{pifont}
\crefname{section}{Sec.}{Secs.}
\Crefname{section}{Section}{Sections}
\Crefname{table}{Table}{Tables}
\crefname{table}{Tab.}{Tabs.}
\Crefname{appendix}{Appendix}{Appendices}
\crefname{appendix}{Appx.}{Apps.}

\usepackage{algorithm}
\usepackage{algorithmic}
\usepackage{amsmath}

\title{Computational Bottlenecks of Training \\ 
Small-scale Large Language Models}

\hypersetup{
  colorlinks,
  linkcolor={red!50!black},
  citecolor={blue!50!black},
  urlcolor={blue!80!black}
}

\author{%
Saleh Ashkboos\thanks{Work done during an internship at Apple.}
\And
Iman Mirzadeh
\And
Keivan Alizadeh
\And
Mohammad Hossein Sekhavat
\And
Moin Nabi
\And
Mehrdad Farajtabar
\And
Fartash Faghri
\AND
\vspace*{-0.7cm}\\
Apple\\
\texttt{saleh.ashkboos@inf.ethz.ch,fartash@apple.com}
}

\DeclareTextFontCommand{\emph}{\em}

\begin{document}

\maketitle

\begin{abstract}

While large language models (LLMs) dominate the AI landscape, Small-scale large Language Models (SLMs) are gaining attention due to cost and efficiency demands from consumers. However, there is limited research on the training behavior and computational requirements of SLMs. In this study, we explore the computational bottlenecks of training SLMs (up to 2B parameters) by examining the effects of various hyperparameters and configurations, including GPU type, batch size, model size, communication protocol, attention type, and the number of GPUs. We assess these factors on popular cloud services using metrics such as \emph{loss per dollar} and \emph{tokens per second}~\footnote{We use average dollar cost ratios of cloud instance types based on publicly available pricing (\cref{appx:hardware_details}).}. Our findings aim to support the broader adoption and optimization of language model training for low-resource AI research institutes.
\end{abstract}

\section{Introduction}



Large Language Models (LLMs) are becoming increasingly popular in various fields due to their performance on a variety of tasks~\cite{brown2020language,openai2023gpt4,devlin2018bert,raffel2020exploring,bommasani2021opportunities}. However, deploying large models widely such as on mobile hardware and edge devices is challenging due to the large memory and compute requirements~\cite{hoefler2021sparsity, hu2021lora, gou2021knowledge}.
These constraints have driven a growing interest in smaller language models (such as $\leq2B$ parameters) as a viable alternative~\citep{zhang2024tinyllama,liu2024mobilellm,xia2023sheared}. Recent work refer to these models as Small-scale large Language Models (SLMs) which can work well in environments where cost-efficiency and resource limitations are of significant concern, as well as on servers where the reduced cost of inference will be a dominant factor to attract and retain customers.

SLMs have demonstrated substantial potential in achieving competitive results despite their smaller size. 
Techniques such as pruning, distillation, and quantization have been employed to enhance their performance~\citep{slicegpt, quarot, distil}, allowing SLMs to perform on par with, and in some cases surpass, much larger models~\cite{bansal2024smaller}.
For example, Gemma-2B outperforms the largest OPT-175B~\citep{opt}, challenging the notion that sheer model size is the primary determinant of effectiveness. In addition to on par accuracy, SLMs can meet consumer demands for fast, efficient, and cost-effective AI without sacrificing task performance, making them increasingly attractive for organizations with limited computational budgets, such as small businesses and academic institutions.

While prior work mostly focused on optimizing SLMs for inference~\cite{kim2023full}, relatively little attention has been paid to their training dynamics. This gap is significant, as the computational and infrastructure demands of training LLMs may not translate to SLMs. Given the diverse range of hardware configurations available on cloud platforms—such as GPU type, batch size, and communication protocols—there is a need for a systematic analysis of how these factors impact the training efficiency of SLMs, particularly when measured in terms of practical metrics such as \emph{loss per dollar} and \emph{tokens per second}. Our findings indicate that for smaller models, more affordable options like A100-40GB GPUs and Distributed Data Parallel (DDP) can be utilized without sacrificing performance. For larger models, advanced configurations, such as A100-80GB and H100-80GB GPUs paired with Flash Attention (FA) and Fully Sharded Data Parallel (FSDP), are necessary to handle larger batch sizes and prevent memory-related issues.

Recent advancements in the field underscore the importance of scaling AI systems not only for state-of-the-art performance but also for practical applications in real-world environments. The emerging trend toward SLMs suggests that a re-evaluation of hardware and computation strategies is essential. The contribution of this paper is to address the need for such  evaluation, providing a systematic study on the computational bottlenecks and cost-efficiency of training SLMs up to 2B parameters on various cloud infrastructure and setups.
We find that
1) FlashAttention is significantly more important for SLMs than LLMs,
2) Expensive hardware, e.g., H100-80GB and A100-80GB, is not necessarily cost effective for SLM training,
3) DDP is the best distributed training scheme for SLMs, and
4) Maximizing GPU memory utilization is not cost-optimal for SLM training.


\section{Metrics}\label{sec:metrics}

Our goal is to find architectures with maximal performance
and minimum cost of training. It is common to measure the cost of training in terms of
wall-clock time, iterations, or tokens. However, these metrics are incomplete
for choosing a sufficient infrastructure within a budget.
We recommend metrics that directly incorporate the dollar cost of hardware.
Specifically, we aim to
maximize the accuracy of the model while minimizing the cost or in other words optimize for \textit{accuracy/dollar}. Prior works have discovered neural scaling laws controlling the relation between accuracy,
loss, and number of seen samples or tokens during training~\citep{kaplan2020scaling}. In this paper,
we focus on  the number of tokens processed during the training (or $\frac{\text{Token}}{\text{Second}}$) for various architectures and measure the cost of processing tokens (or $\frac{\text{Token}}{\text{Dollar}}$). Given
$\frac{\text{Token}}{\text{Sec}}$ measurements and $\frac{\text{Loss}}{\text{Token}}$ derived from scaling laws, we get
\begin{equation}\label{eq:loss_per_dollar}
    \frac{\text{Loss}}{\text{Dollar}}=\frac{\text{Loss}}{\text{Token}} \times\underbrace{ \frac{\text{Token}}{\text{Second}}\times inv(\frac{\text{Dollar}}{\text{Second}})}_{\text{Our Metric}},
\end{equation}
where, $\frac{\text{Dollar}}{\text{Second}}$ is the cost of hardware and infrastructure which is extracted by averaging publicly available prices from various cloud providers for each hardware configuration (See \cref{appx:hardware_details}). 
The result of our analysis provides $\frac{\text{Token}}{\text{Dollar}}$ laws that combined with loss scaling laws can be used to find the minimal cost to a target loss.
Other metrics of interest can be CPU/GPU utilization and Memory bandwidth usage that we leave for future work. We will report $\frac{\text{Token}}{\text{Dollar}}$ in various setups below where a value of 1k means training for 1k tokens costs \$1.

\section{Model and Parameters}\label{sec:model_params}

We focus on LLaMa architectures~\cite{llama2, llama3} as they are one of the most popular architectures in recent public LLMs and SLMs~\citep{phi3, mistral}. The smallest LLaMa-2/3 have 7/8B parameters which is still too large for most mobile hardware. We extract the number of decoder blocks and parameters of our models by fitting a curve over the LLaMa models and use it for defining our models (see \cref{fig:scaling_law_model_size} in \cref{appx:model_details}). We evaluate four different model sizes with 100M, 500M, 1B, and 2B parameters.  \textit{Notably, we maximize over all configuration parameters not shown in the x-axis or legend of a figure.} That is, we run a large grid search over all combinations of configuration parameters listed below and each point in each plot is the best configuration given all parameters specified in the plot. This way, we find the the optimal $\frac{\text{Token}}{\text{Dollar}}$ and assume one can tune optimization hyperparameters such as learning rate to achieve the optimal convergence with the hardware-optimal configurations. We present details of these derivative models in \cref{appx:model_details}.
Next, we define the configuration parameters:

\vspace*{-2mm}
\begin{itemize}[leftmargin=*]
\itemsep0em
    \item \textbf{GPU Types}: We evaluate the usage of three NVIDIA GPU types:  A100-40GB,  A100-80GB, and H100-80GB. We use BFloat16 data types in all GPUs.
    \item \textbf{GPU Numbers and Communication}: We study three main training configurations for each GPU type including single-node-single-GPU (1 GPU), single-node-multi-GPU (2, 4, and 8 GPUs), and multi-node-multi-GPU (16, 32, and 64 GPUs) settings. When we use more than a single GPU, we evaluate Distributed Data-Parallel (DDP) and Fully Sharded Data Parallel \cite{fsdp} (FSDP)
    for communication. For FSDP, we study two sharding policies: 1) \texttt{full} sharding where we shard all gradients, optimizer states, and weights, and 2) \texttt{grad\_op} sharding where we shard only gradients and optimizer states (but keep the weights unsharded). We use RDMA/EFA.
    \item \textbf{Number of Samples}: We evaluate various number of samples fit into a single GPU during the training. We fix the sequence length to 1028 and iterate over the batch-size we fit into a single device. As we cannot fit 128 samples into a single GPU memory even in our smallest (100M) model, we study the per-device batch-size of 4, 8, 16, 32, and 64. We do not use gradient accumulation.
    \item \textbf{Flash Attention}: We study the affect of using FlashAttention \cite{flashattention} for the attention block.
    
\end{itemize}


\section{Experimental Results}

In this section, we present results on  A100-40GB, A100-80GB, and H100-80GB. We implement our models in HuggingFace~\cite{hf_transformers} and run our experiments using PyTorch~\cite{pytorch} without any additional frameworks and use 1024 sequence length. We provide details of the runtime configuration in \cref{appx:runtime_details}. For each setup, we run our experiments at least 3 times, each with 10 training steps, and then report the average and error bars. We form our findings into research questions and answers.

\textbf{Q1: How important is to use FlashAttention during the SLM training?}

\begin{figure}[t!]
\centering\includegraphics[width=0.99\textwidth]{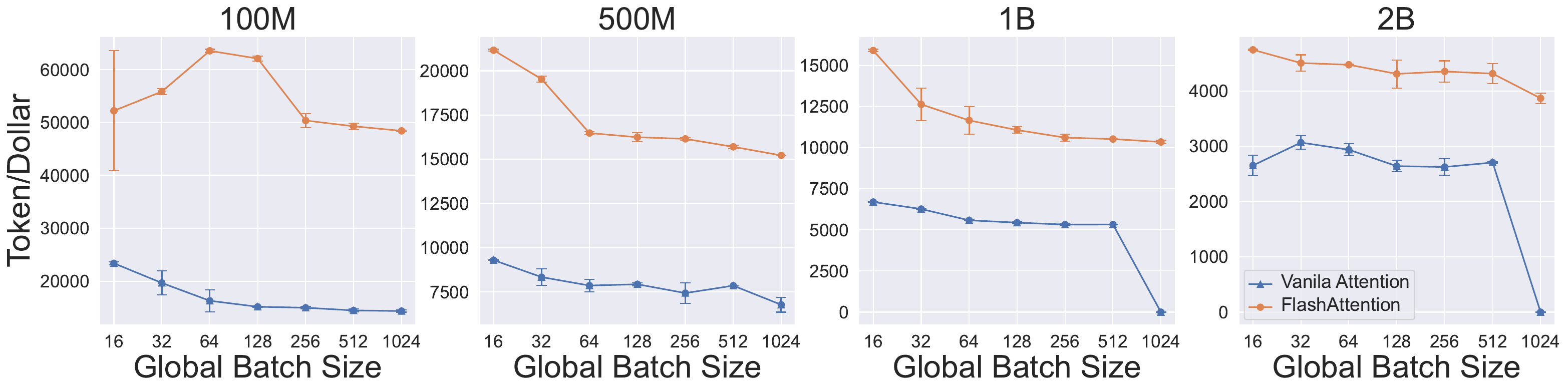}
\vspace{-5pt}
\caption{\textbf{FlashAttention is more cost-efficient for smaller models.} Maximum $\text{Token}/\text{Dollar}$ across GPU-types, GPU-number, and communication type when we use FlashAttention. FlashAttention shows a significant $\text{Token}/\text{Dollar}$ improvement over vanilla attention in smaller models and batch sizes. OOM runs are shown with 0.
$\text{Token}/\text{Dollar}=$1k means training for 1k tokens costs \$1.}
\label{fig:flashAttn}
\vspace{-7pt}
\end{figure}

\Cref{fig:flashAttn} compares the use of FlashAttention2~\citep{flashattention} against vanilla attention for different global batch-sizes. First, we can see that FlashAttention  (FA) significantly increases our $\text{Token}/\text{Dollar}$ in SLMs.
Notably, FA improves $\text{Token}/\text{Dollar}$ more significantly for smaller models
as the cost of attention is quadratic in context length and dominates when the hidden dimension shrinks.
Such SLMs enter a data-bound regime where the data movement (CPU/GPU as well as GPU/GPU) becomes the main bottleneck. Finally, we can see that for larger models (1B and 2B), FA enables training  of larger batch-sizes (1024) while vanilla attention results in out of memory error (OOM).

\textbf{Q2: Given a fixed number of GPUs, what is the best GPU type for training an SLM?}

\begin{figure}[t!]
\centering\includegraphics[width=0.99\textwidth]{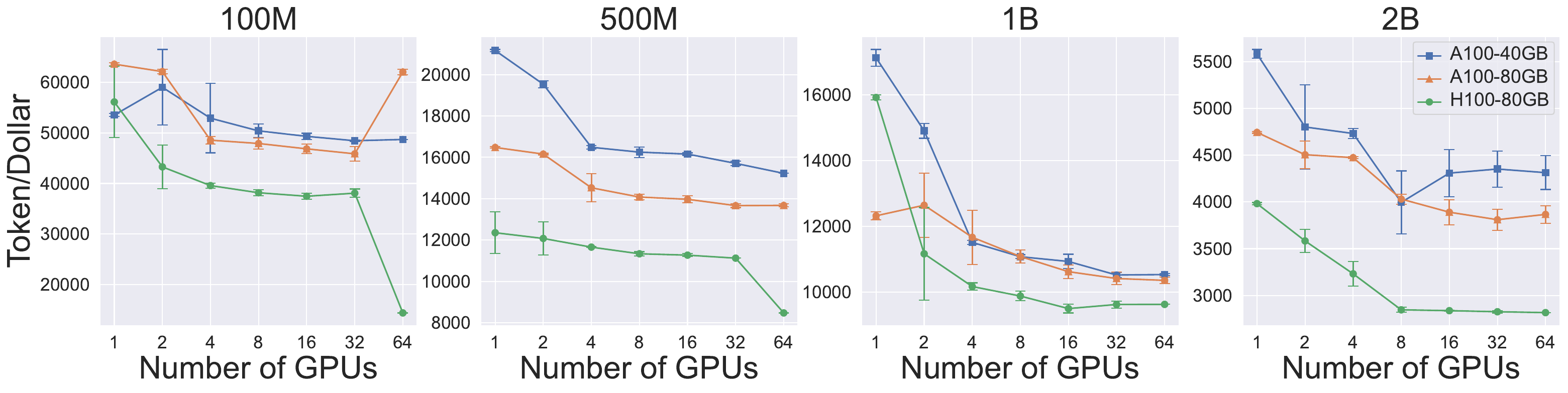}
\vspace{-5pt}
\caption{\textbf{H100 GPUs are not cost-efficient for training SLMs.} Maximum $\text{Token}/\text{Dollar}$ for different GPU-type across batch-size and communication types. We use FlashAttention.}
\label{fig:gpu_type} 
\vspace{-1em}
\end{figure}

\Cref{fig:gpu_type} shows the result of training our models using different GPU types: A100-40GB, and A100-80GB. Although we cannot see a consistent pattern for all models, we can see that A100-80GB GPU is a better choice when we use a large number of GPUs (32) to train larger models (1B and 2B). In such cases, A100-80GB can be used for training larger batch-sizes, while in smaller models, we can use 40GB GPU with cheaper price (see \cref{appx:hardware_details} for Hadrware prices).

\textbf{Q3: What is the best communication scheme for training SLMs for different number of nodes?}

\begin{figure}[h!]
\centering\includegraphics[width=0.99\textwidth]{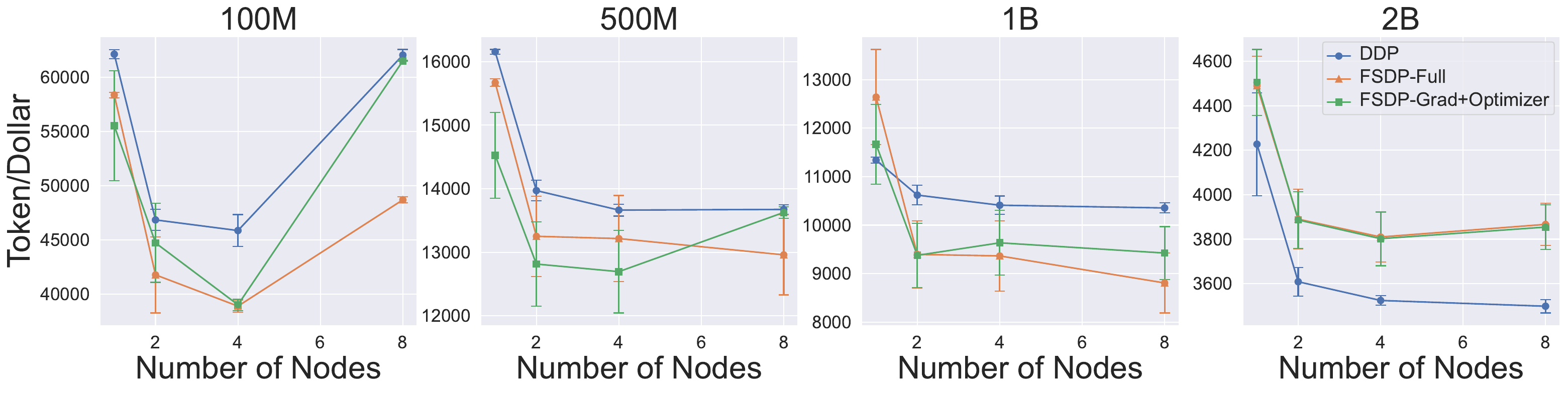}
\vspace{-5pt}
\caption{\textbf{DDP is the best scheme for training SLMs.} Maximum $\text{Token}/\text{Dollar}$ for different GPU-nodes on A100-80GB across different batch-sizes. We use FlashAttention in our models. for a single node, we use 2, 4, and 8 GPUs while for 2 and 4 nodes we use 16 and 32 GPUs respectively.}
\label{fig:comm_gpunodes_80gb} 
\vspace{-1em}
\end{figure}

Next, we study the role of using different parallelization schemes for SLM training. To this end, we study the use of Distributed Data Parallel (DDP), Fully Sharded Data Parallel  with full sharding policy (FSDP-Full), and Fully Sharded Data Parallel  with sharding gradients and optimizer states (FSDP-Grad+Optimizer). \Cref{fig:comm_gpunodes_80gb} shows the result of training our models with different parallelization schemes on A100-80GB GPU. Our results show that for smaller models, DDP is a better choice due to the less communication volume. However, for largest model (2B), FSDP outperforms DDP as we can train larger batch-sizes (see Q4). Finally, we observe that FSDP-Grad+Optimizer outperforms FSDP-Full due to the lower communication overhead.

\begin{figure}[h!]
\centering\includegraphics[width=0.99\textwidth]{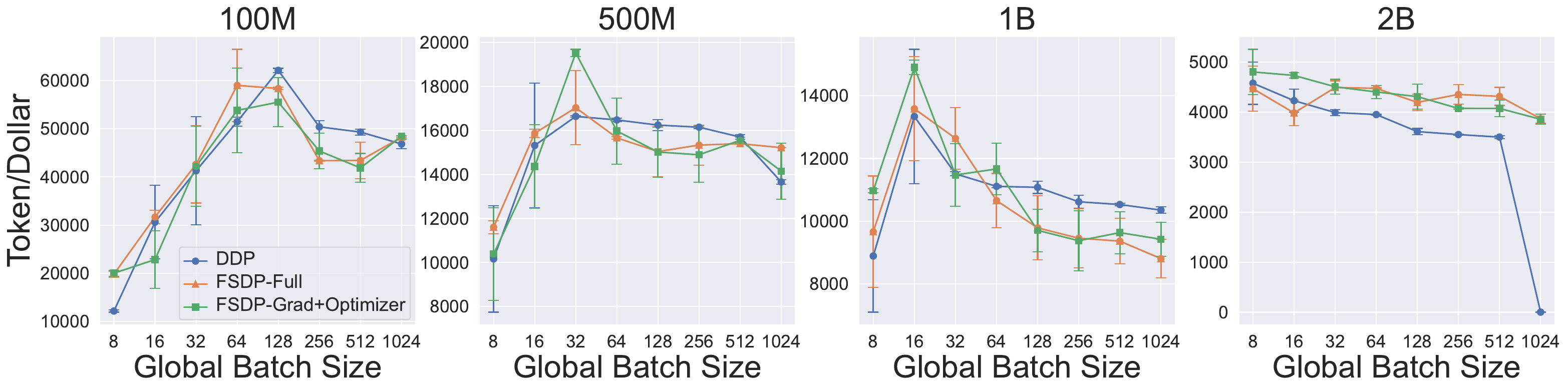}
\vspace{-5pt}
\caption{\textbf{For SLMs increasing global batch size saturates cost-efficiency before GPU memory is fully utilized.} Maximum $\text{Token}/\text{Dollar}$ for different global batch-sizes across across different GPU-types, GPU numbers, and various per-device batch-sizes. We use FlashAttention in our models.}
\label{fig:comm_bsz} 
\vspace{-1em}
\end{figure}

\textbf{Q4: What is the best communication scheme for training SLMs for different global batch-sizes?}
\cref{fig:comm_bsz} shows the result of training SLM with various global batch-sizes using DDP, FSDP-Full, and FSDP-Grad+Optimizer for various per-device batch sizes. We cannot see a substantial difference in small batch sizes in our experiments. However, similar to Q3, FSDP always outperforms DDP for largest model (2B) and batch-sizes. In addition, FSDP enable training larger global batch-sizes (512) for larger models (2B) compared to DDP (which results in OOM).

\vspace{-0.5em}
\section{Summary and Conclusion}
In this study, we examined the computational bottlenecks of training Small-scale Language Models (SLMs) up to 2B parameters, focusing on the impact of hyperparameters and hardware configurations. 
Our findings highlight the importance of Flash Attention (FA) for smaller models and batch sizes, where data movement is the primary bottleneck. FA also enables training larger models (1B and 2B) with batch sizes up to 512, avoiding out-of-memory (OOM) errors common with vanilla attention. 
Additionally, we found that A100-80GB GPUs are optimal for training larger models with many GPUs, while the more cost-effective A100-40GB works well for smaller models.
In terms of distributed training, DDP is more efficient for smaller models, but FSDP outperforms it for larger models, particularly when training large batch sizes.
These insights provide practical guidance for optimizing SLM training by offering clear strategies for selecting the most efficient hardware and parallelization methods based on model size.
\newpage








\bibliography{References}
\bibliographystyle{plainnat}

\newpage
\appendix
\label{Appendix}
\section{Hardware Costs: Details}\label{appx:hardware_details}

To get reasonable (and practical) GPU price, we extracted the price of allocating different GPU types from two providers: \textbf{Lambda Labs}\footnote{\url{http://lambdalabs.com}}, and \textbf{Google Cloud Platform (GCP)}\footnote{\url{https://cloud.google.com}}. We then normalized and average over them to use the proportional ratio in our experiments. Table \ref{tab:prices} shows the details of our hardware prices.

\begin{table}[H]
 
    \centering
    \begin{tabular}{|c|c|c|c|}
    
\toprule
 GPU Type  & Lambda Labs & GCP & Avg.\\
\midrule
A100-40GB & 1.00 & 1.00  &  1.00 \\
A100-80GB & 1.39 & 1.54  &  1.46 \\
H100-80GB & 2.32 & 2.73  &  2.53 \\
\bottomrule
\end{tabular}
\vspace{0.5em}
    \caption{
    Normalized prices for a single GPU allocation (for an hour) on \textbf{Lambda Labs} and \textbf{Google Cloud Platform (GCP)}. We use averged values in our experiments.
    }
    \label{tab:prices}
\end{table}

\section{Model Parameters: Details}\label{appx:model_details}

\begin{figure}[h!]
\centering\includegraphics[width=0.96\textwidth]{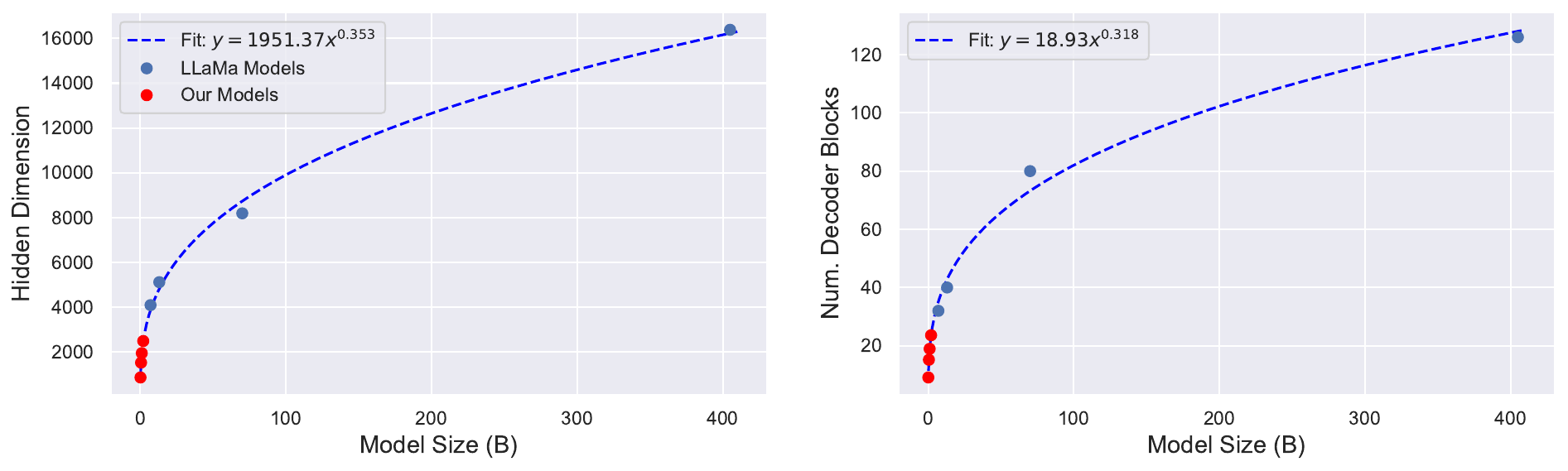}
\caption{\label{fig:scaling_law_model_size} Curve fitting to extract our SLM model sizes using LLaMa-2/3 models. \textbf{Left}: Hidden dimension extraction. \textbf{Right}: Number of decoder blocks.}
\vspace{-1em}
\end{figure}

Table \ref{tab:model_details} shows the architecture details of the models we used in this paper.

\begin{table}[H]
 
    \centering
    \begin{tabular}{|c|c|c|c|}
    
\toprule
Model  & \#L & \#D & \#Param. \\
\midrule
100M & 9  &  864  &    130,569,920 \\
500M &  15 &  1536  &  526,550,016  \\
1B &  19 &   1920 &   970,458,240 \\
2B &  24 &   2496  &  1,969,027,008  \\
\bottomrule
\end{tabular}
\vspace{1em}
    \caption{
    Model details in this study. We use LLaMa-style architectures and show the number of decoder layers (\textbf{\#L}), the dimension of the model (\textbf{\#D}), and the exact number of parameters (\textbf{\#Param.}).
    }
    \label{tab:model_details}
\end{table}

\section{Runtime Configuration: Details}\label{appx:runtime_details}
Table \ref{tab:version_details} presents the PyTorch version and its dependencies used in the runtime environment of this study.

\begin{table}[H]
    \centering
    \begin{tabular}{|c|c|}

\toprule
Dependency & Version \\
\midrule
PyTorch & 2.3.0 \\
NCCL & 2.20.5 \\
CUDA & 12.2.2 \\
Fabric interface provider & 119.0 \\ 
Libfabric API & 1.22.0 \\
Python & 3.10.13 \\
Ubuntu & 20.04 \\
\bottomrule
\end{tabular}
\vspace{1em}
    \caption{
    Dependency versions.
    }
    \label{tab:version_details}
\end{table}

\end{document}